\documentclass[letterpaper, 10 pt, journal, twoside]{IEEEtran}
\usepackage{epsfig} % for postscript graphics files
\usepackage{mathptmx} % assumes new font selection scheme installed
\usepackage{times} % assumes new font selection scheme installed
\usepackage{amsmath} % assumes amsmath package installed
\usepackage{amssymb}  % assumes amsmath package installed
\usepackage{hyperref}
\usepackage{cite}
\usepackage[cal=cm]{mathalpha}
\usepackage{amssymb,amsfonts}
\usepackage{graphicx}
\usepackage{xcolor}

\usepackage{balance}
\DeclareMathOperator*{\argmax}{\arg\!\max}

\begin{document}
\title{MPPI-based Informative Trajectory Planning for \\ Search and Capture of Drifting Targets with ASVs}
\author{Sanjeev Ramkumar Sudha, Marija Popovi{\'c}, and Erlend M. Coates% <-this % stops a space
\thanks{Manuscript received: June 10, 2026; Revised: September 6, 2026; Accepted: October 15, 2026. This paper was recommended for publication by Editor Giuseppe Loianno upon evaluation of the Associate Editor and Reviewers' comments. Corresponding author: Sanjeev Ramkumar Sudha.}
\thanks{Sanjeev Ramkumar Sudha and Erlend M. Coates are with the Norwegian University of Science and Technology (NTNU), 6009 Ålesund, Norway. {\tt\small\{sanjeev.k.r.sudha, erlend.coates\}@ntnu.no}. Marija Popovi{\'c} is with Delft University of Technology, 2629 Delft, The Netherlands. {\tt\small m.popovic@tudelft.nl}.}% <-this % stops a space
\thanks{Digital Object Identifier (DOI): see top of this page.}
}% note the % following the last \IEEEmembership and also \thanks - 

\markboth{IEEE Robotics and Automation Letters. Preprint Version. Accepted October, 2026}
{Ramkumar Sudha \MakeLowercase{\textit{et al.}}: MPPI-based Informative Planning for Search and Capture} 




\maketitle

\begin{abstract}
Autonomous surface vehicles (ASVs) offer an efficient solution for environmental cleanup as well as search and rescue operations in open waters. Targets in these settings drift continuously, so efficient search must balance exploration of unobserved regions with tracking of known targets. 
However, most target tracking and pursuit scenarios consider simple guidance behaviours and short-term predictions for decision-making. 
In this letter, we address the problem of search and capture of multiple drifting targets, such as litter, in marine environments, using an adaptive informative planning approach. A key aspect of our strategy is a spatiotemporal informative planning method based on model predictive path integral (MPPI) control, a sampling-based model predictive control method. Our planner balances search and tracking objectives with a cost design that couples spatiotemporal target drift predictions and exploration objectives.
The planner directly generates kinematic-level commands by optimising continuous trajectories over long horizons. 
In the interception stage, we use pure pursuit guidance for physical capture of the targets.
Simulations show that our planner captures up to 18\% more targets than the strongest baseline, proving its efficiency for environmental cleanup. Finally, we validate our approach in field trials with an ASV.
\end{abstract}

% \begin{IEEEkeywords}
% IEEE, IEEEtran, journal, \LaTeX, paper, template.
% \end{IEEEkeywords}

% \fi
\IEEEpeerreviewmaketitle

%%%%%%%%%%%%%%%%%%%%%%%%%%%%%%%%%%%%%%%%%%%%%%%%%%%%%%%%%%%%%%%%%%%%%%%%%%%%%%%%

\section{Introduction}

\IEEEPARstart{M}{aritime} search and capture operations are critical for both search and rescue and environmental cleanup. 
% About 8–12 million tonnes of plastic enter the world’s oceans every year.
% Floating pollutants, such as plastic litter, can sink and degrade into toxic microplastics if allowed to persist, proving lethal to marine life \cite{jambeck2015plastic}, making timely mapping and cleanup of pollutants essential.
Floating pollutants, such as plastic litter and oil spills, are lethal to marine life, making timely mapping and cleanup of such pollutants essential.
% At least one in three marine mammal species becomes entangled in litter \cite{kuhn2015deleterious}.
Autonomous robots for environmental monitoring \cite{dunbabin2012robots} offer a safer and more cost-effective alternative compared to traditional approaches such as manual surveys \cite{murphy2017disaster} and static sensor networks \cite{lanzolla2021wireless}. However, most existing robot monitoring approaches rely on passive sensing \cite{galceran2013survey} and do not actively adapt sensing toward regions of interest. While adaptive planning techniques \cite{hollinger2014sampling, bircher2016receding, charrow2015information, asgharivaskasi2022active} direct sensing toward regions of interest during the mission, this remains challenging in dynamic environments.

This letter addresses the problem of search and capture of floating litter using an autonomous surface vehicle (ASV). Freely floating objects in open waters drift under the influence of wind and currents. In \textit{a priori} unknown environments, the ASV has to simultaneously explore unobserved regions to detect new targets and track previously detected targets to intercept them before they drift away, creating an exploration-exploitation trade-off. Efficient monitoring in such scenarios requires planning vehicle trajectories over long horizons that reason over future target positions. Precise guidance and control, coupled with perception, are required to physically intercept drifting targets.

\begin{figure}[!t]
    \centering
    \includegraphics[width=0.85\linewidth]{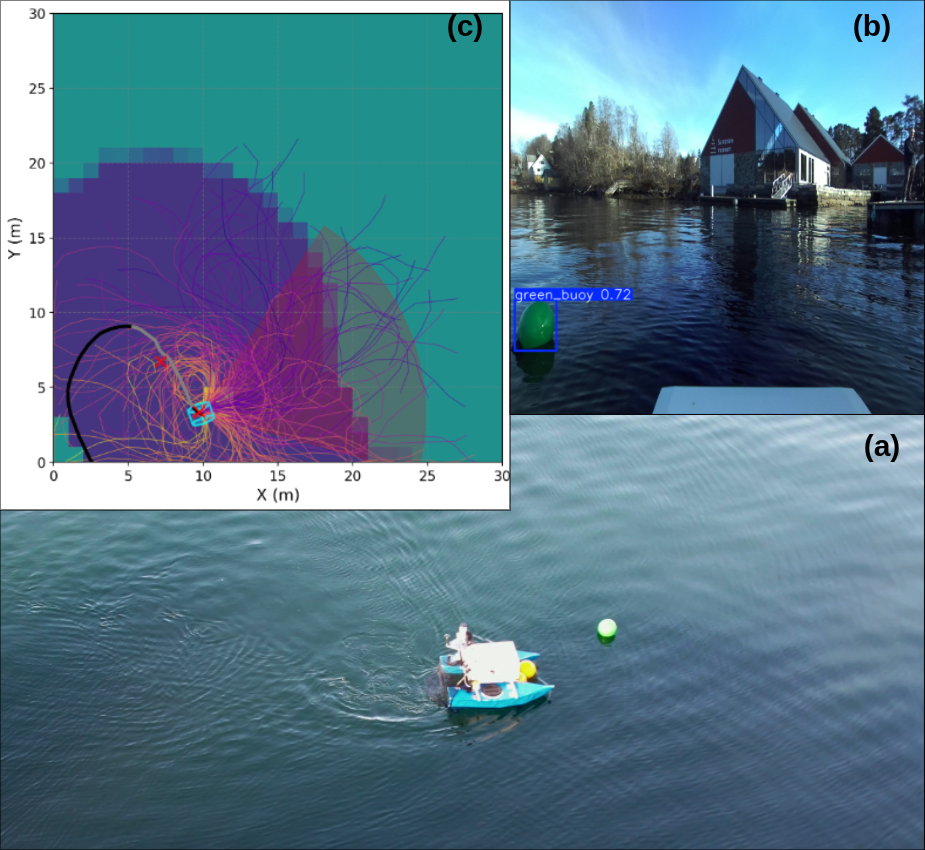}
    \caption{(a) An autonomous surface vehicle (ASV) during field tests for search and capture. (b) A stereo camera is used for the detection and localisation of targets of interest. (c) We use a dynamic occupancy grid for mapping. A key aspect of our approach is the proposed informative planning method, based on model predictive path integral (MPPI) control. 
    % Our cost design and a modified sampling distribution for the MPPI enables ..
    }
    \label{fig:fig1}
\end{figure}

Target search and tracking is a widely researched problem in robotics \cite{robin2016multi}, where targets of interest are detected and localised. 
% Static target search with mobile robots involves searching for targets of interest in unknown environments \cite{meera2019obstacle}. 
Persistent monitoring is a related problem that generally involves estimating the positions of non-adversarial and non-cooperative mobile targets, by periodically re-detecting them \cite{wang2023spatio, yuan2025tracking}. 
% Pursuit and evasion is another related body of research that considers targets that actively try to evade pursuers, which try to capture them \cite{chung2011search}.
% While being well-studied, the decision-making in such problems is generally limited to short-term predictions and simple guidance behaviours such as chasing a target until a criterion is met.
However, decision-making in such scenarios is typically limited to greedy guidance behaviours that optimise a single objective over short horizons, without jointly reasoning over exploration and future target distributions.

The \textit{informative planning} problem deals with planning a sequence of actions that optimise an information-theoretic objective. Sampling-based planners are commonly used due to their efficiency in high-dimensional action spaces \cite{bircher2016receding, hollinger2014sampling}. Some studies also explore the use of trajectory optimisation techniques, such as model predictive control (MPC), for active sensing and informative planning \cite{charrow2015information, asgharivaskasi2022active, falanga2018pampc, bostrom2018global}. They usually offer better theoretical guarantees but generally require the objectives to be differentiable and are solved online with gradient-based solvers or non-smooth optimisation tools, which can be computationally expensive for planning in real-time. 
% In the case of dynamic and unknown environments, incrementally building a map and predicting changes to the map is required. Adaptive planning is necessary in such environments, which requires estimating potential rewards along possible trajectories by predicting both spatial and temporal variations.
Planning in dynamic environments is challenging, as it requires actively updating and predicting changes to the map while planning trajectories that estimate the informativeness of possible trajectories by accounting for both spatial and temporal variations in the map.

% The main contribution of this paper is a spatiotemporal informative planning based on  model predictive path integral (MPPI) control to tackle the problem of search and capture of moving targets with an ASV, such as litter in marine environments. Our planner switches between the global informative planner for exploration and target tracking and a pure pursuit guidance scheme for target interception. We develop a new spatiotemporal informative planning approach based on model predictive path integral (MPPI) control, a sampling-based model predictive control (MPC) approach. This allows us to output velocity-level commands directly while considering longer prediction horizons, without requiring the cost formulation to be convex or differentiable over the control and state space.Our cost design enables planning trajectories that effectively trade off multiple objectives, such as searching for targets (exploration) and intercepting detected targets (exploitation) while adhering to collision and dynamic constraints. When the vehicle is in close vicinity of a target, we switch to a different controller based on pure pursuit guidance to ensure successful capture of the target. 

% We develop a new spatiotemporal informative planning approach based on model predictive path integral (MPPI) control, a sampling-based model predictive control (MPC) approach.
The main contribution of this paper is a spatiotemporal informative planning approach based on model predictive path integral (MPPI) control to tackle the problem of search and capture of drifting targets such as litter in marine environments, with an ASV. 
% Our planner enables efficient search and capture by coupling conflicting objectives, such as searching for targets (exploration) and intercepting moving targets (exploitation), while adhering to collision and trajectory feasibility constraints. 
The planner cost design incorporates spatiotemporal target drift predictions into the objective, combining them with exploration and trajectory feasibility objectives, while planning in continuous action space. 
This allows us to output velocity-level commands directly while considering longer planning horizons.
% without requiring the cost formulation to be convex or differentiable over the control and state space.
Our adaptive planning framework switches between the global informative planner for exploration and target tracking and a pure pursuit guidance scheme for target interception. 
% When the vehicle is in close vicinity of a target, we switch to a different controller based on pure pursuit guidance to ensure successful capture of the target. 

%% VERSION 1
\noindent To summarise, our contributions are as follows:
\begin{itemize}
    \item A spatiotemporal informative planning approach based on MPPI with a cost design that reasons over predicted target drift to jointly optimise exploration and target interception over continuous trajectories. 
    % Our planner outperforms adaptive and non-adaptive planning baselines in simulation, and ablation studies validate the effectiveness of each cost component.
    \item A hybrid planning framework combining an MPPI-based informative planner and a pure pursuit guidance controller for search and capture of drifting targets with an ASV.
    \item Evaluations in simulation show that our planner outperforms baselines, and ablation studies further validate the cost design. Field experiments with an ASV demonstrate the framework's validity in real-world scenarios. We also open source our code at \href{https://github.com/sanjeevrs2000/mppi-search-and-capture}{github.com/sanjeevrs2000/mppi-search-and-capture}.
    % \href{https://anonymous.4open.science/r/mppi-search-and-capture-4577}{\footnotemark}.

% \footnotetext{\url{anonymous.4open.science/r/mppi-search-and-capture-4577}}

\end{itemize}

%% VERSION 2
% \textcolor{red}{
% .. Our method is evaluated against other planning baselines in realistic simulations in Gazebo. We also perform ablation studies to evaluate the effectiveness of our cost design. Additionally, field experiments are conducted with an ASV, validating our framework for real-world monitoring scenarios.
% To summarise our contributions, we claim the following:
% \begin{itemize}
%     \item The proposed trajectory optimisation planner in continuous space achieves better task completion rates than adaptive and non-adaptive planning baselines. 
%     \item Ablation study of our reward design shows that our reward design outperforms purely greedy and exploratory planning behaviours.
%     % \item We validate our framework in field experiments 
%     % \item We show that utilising trajectory optimisation with our proposed multiobjective utility for global exploration outperforms myopic explorative or tracking baselines.
% \end{itemize}
% }

\section{Related Work}

% analogous applications - target search, pursuit evasion, spatiotemporal mapping etc

% Search and tracking of dynamic targets in unknown environments is a widely researched problem in the literature. In such scenarios, the agent needs to search for targets in an unknown environment (exploration) and track known targets to improve their state estimates (exploitation). Some studies consider persistent monitoring of multiple targets with a single robot, where the goal is to plan trajectories for the robot to redetect targets frequently \cite{sanjeev2026informative, wang2023spatio}. Many studies also consider multirobot cooperation for monitoring multiple targets \cite{wang2022marine, jeong2026multi, dames2017detecting, hung2025target}. Pursuit and evasion is another related problem where the targets actively try to evade the pursuing robot \cite{chung2011search}. In these studies, the decision-making is often limited to greedy behaviours such as pursuing a single target, without jointly optimising the exploration and exploitation objectives over continuous trajectories. While these studies are closely related to our problem, we specifically consider the pursuit and capture of non-evasive targets in dynamic marine environments, where the robot must balance exploration of unknown regions with interception of detected targets.

Persistent monitoring and tracking multiple targets with a single robot has been studied using spatiotemporal attention mechanisms \cite{wang2023spatio} and hierarchical planning under motion uncertainty \cite{yuan2025tracking}, where the goal is to plan trajectories that maintain accurate state estimates of non-cooperative targets. Multi-robot approaches have also been explored for cooperative target monitoring \cite{wang2022marine, jeong2026multi, dames2017detecting, hung2025target}. Plume monitoring and source seeking are analogous problems \cite{du2021multi, hajieghrary2017information}, where partially observable plumes are localised with concentration gradients, whereas in our case the targets are directly observable. Pursuit and evasion is another related problem where the targets actively try to evade the pursuer \cite{chung2011search}. However, the decision-making is often limited to greedy behaviours such as pursuing a single target, without jointly optimising exploration and pursuit over continuous trajectories. Our work differs in that we specifically consider search and physical capture of non-evasive drifting targets in marine environments, where the robot must simultaneously discover new targets and intercept detected ones before they drift out of reach. While we focus on floating litter, the same formulation applies to closely related applications such as maritime search and rescue.

% In contrast to these studies, our approach maintains a global occupancy map of target positions, and the proposed MPPI-based planner plans trajectories by directly considering both the exploration and exploitation objectives.
% Target tracking approaches such as \cite{jeong2026multi, sanjeev2026informative} focus on reducing uncertainty in target positions or maintaining accurate tracks of targets. But they do not consider any interaction with the targets. Some studies consider pursuit and evasion where the targets are actively trying to evade from a pursuer. While these approaches are similar to our problem, we specifically consider pursuit and capture of drifting targets in dynamic marine environments.

% cleanup/tracking related applications
Search and capture of floating targets with multiple robots has been investigated by Barrionuevo et al. \cite{barrionuevo2025optimizing} and Wang et al. \cite{wang2022marine}. However, these approaches are limited to simplified simulation environments and treat capture as a guaranteed event upon close proximity, without addressing the guidance and control challenges of physically intercepting drifting targets. Similar to our approach, Batista et al. \cite{batista2026simtoreal} study search and capture with a single ASV, but their planning is limited to a passive lawnmower pattern with greedy target chasing, and the dynamic nature of the targets is not explicitly considered. In contrast, our proposed adaptive informative planning approach reasons over predicted target drift in a continuous action space.

Sampling-based planners are widely used for informative planning due to their efficiency in high-dimensional action spaces \cite{hollinger2014sampling, bircher2016receding}. Trajectory optimisation methods, such as MPC, have also been applied to informative planning and active sensing \cite{charrow2015information, falanga2018pampc, lodel2022viewpoint}, with the advantage of explicitly reasoning over system dynamics and directly outputting lower-level control commands. However, these methods typically require a smooth cost landscape over the state and control space, and solve the optimisation online with gradient-based solvers or resort to non-smooth optimisation at added cost, which is computationally expensive for real-time deployment. Receding horizon planners with a discrete set of control actions have been studied for spatiotemporal informative planning \cite{chen2019multi, sanjeev2026informative}, but discretisation of the action space limits trajectory diversity and the ability to jointly reason over multiple competing objectives in continuous space. Our approach addresses these limitations by building on MPPI, a sampling-based MPC method that handles non-differentiable, non-smooth cost landscapes while planning over continuous action spaces.

% mppi and closely related applications
% Model predictive path integral (MPPI) \cite{williams2017information} is a sampling-based MPC approach that computes approximate optimal control actions by evaluating costs over many parallel Monte-Carlo rollouts.
MPPI \cite{williams2017information} is a sampling-based MPC approach that computes an approximate solution to the optimal control problem via importance-weighted Monte Carlo rollouts.
% without requiring convexity or differentiability of the cost function. 
Streichenberg et al. \cite{streichenberg2023multi} employ MPPI for cooperative motion planning for ASVs in urban canal environments, and Zhai et al. \cite{zhai2026pamppi} apply it to agile perception-aware quadrotor navigation in unknown environments, demonstrating its effectiveness for real-time navigation with complex objectives. Motivated by these works, we develop an MPPI-based spatiotemporal informative planner for search and capture in marine environments. Our planner takes into account target drift predictions and couples them with search and trajectory feasibility objectives, enabling efficient target search and capture for environmental cleanup. 
% Our multiobjective cost design jointly optimises exploration, target interception, and trajectory feasibility, reasoning over predicted future target distributions.

% However, existing MPPI formulations have not been applied to spatiotemporal informative planning with multiobjective cost designs that jointly reason over exploration and predicted target distributions. Motivated by this gap, we develop an MPPI-based informative planner that addresses these challenges for search and capture in dynamic marine environment
% These studies demonstrate the effectiveness of MPPI for real-time navigation and collision avoidance tasks. 
% While these studies focus on navigation and collision avoidance objectives, we consider dynamic environments where target positions evolve over time. Our multiobjective cost design that jointly optimises exploration, target interception, and trajectory feasibility, reasoning over predicted future target distributions.

\section{Preliminaries}

\begin{figure*}[!htbp]
    \centering
    \includegraphics[width=0.8\linewidth]{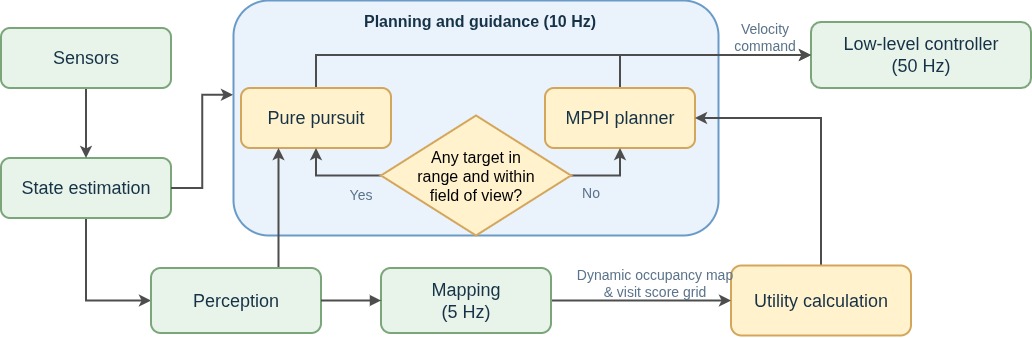}
    \caption{Our proposed framework for search and capture of drifting targets in marine environments. A stereo camera detects and estimates target positions. We use a dynamic occupancy map to keep track of target positions. The key components of our hybrid planning approach include an informative trajectory planning formulation based on MPPI that enables efficient target search and capture. For target interception, we use pure pursuit guidance.}
    \label{fig:flow_diag}
\end{figure*}

\subsection{Problem Statement}

We consider search and capture of moving targets with an ASV in \textit{a priori} unknown, dynamic environments, where targets drift freely under wind and currents. Our aim is to maximise the number of successful captures within a limited mission time. We frame this as an informative planning problem, where the goal is to find a trajectory $\mathcal{T}$ from the set of feasible trajectories $\Psi$ that maximises an information-theoretic objective $I(\mathcal{T})$ subject to a resource constraint $B$:
\begin{equation}
    \mathcal{T}^* = \argmax_{\mathcal{T} \in \Psi}\; I(\mathcal{T}) \quad \text{s.t.} \quad \mathcal{C}(\mathcal{T}) \leq B,
\end{equation}
where $\mathcal{C}(\mathcal{T})$ is the resource value associated with executing trajectory $\mathcal{T}$. 
In time-varying environments, predicting the spatial and temporal variations in target positions is necessary to plan informative trajectories. % -> remove this line?
We use an MPPI-based planning approach (Sec. \ref{sec:planning}) to optimise trajectories in continuous action space, where we minimise a planning cost $J$, equivalent to maximising $I$, over a finite receding horizon, with the constraint $B$ enforced as the total mission duration. %Our planning
% In our approach, we plan trajectories in continuous action space using a planning utility $I(\mathcal{T})$ that considers exploration and exploitation objectives, trajectory feasibility, and collision avoidance, as described later in Sec. \ref{sec:approach}.

\subsection{Model Predictive Path Integral Control}

MPC solves the optimal control problem over a finite horizon $T$ at each time step. Given the current state $\mathbf{x}_0$, the goal is to find a control sequence $\mathbf{U}^* = \{\mathbf{u}_0^*, \ldots, \mathbf{u}_{T-1}^*\}$ subject to dynamics $\mathbf{x}_{t+1} = f(\mathbf{x}_t, \mathbf{u}_t)$, that minimises the total cost:
\begin{equation}
    J(\mathbf{U}) = \phi(\mathbf{x}_T) + \sum_{t=0}^{T-1} \gamma^t C(\mathbf{x}_t, \mathbf{u}_t),
\label{eq:optimal_control_objective}
\end{equation}
\noindent where $\gamma$ is the discount factor, usually set to $1$, $C(\mathbf{x}_t, \mathbf{u}_t)$ is the stage cost and $\phi(\mathbf{x}_T)$ is the terminal cost. Only the first control input $\mathbf{u}_0^*$ is applied, before the optimisation is solved again in the next time step.
% Conventional MPC solvers rely on gradient-based optimisation, which requires the cost function and dynamics to be convex or differentiable, limiting real-time deployment with complex objectives.

MPPI \cite{williams2017information} uses importance-weighted Monte Carlo sampling to find an approximate solution to the optimal control problem. The objective can be an arbitrary non-differentiable or non-smooth cost over the control space. While nonlinear MPC and non-smooth optimisation methods can handle non-convex costs, they typically increase computational complexity and runtime, whereas MPPI evaluates costs by sampling. Given a nominal control sequence $\mathbf{U}$, trajectories are sampled by perturbing each control input with added noise $\boldsymbol{\epsilon}_t^{(k)} \sim \mathcal{N}(\mathbf{0}, \Sigma)$. The total cost of trajectory $k$ is:
\begin{equation}
    S^{(k)} = \phi\!\left(\mathbf{x}_T^{(k)}\right) + \sum_{t=0}^{T-1} \gamma^t\left[ C\!\left(\mathbf{x}_t^{(k)}, \tilde{\mathbf{u}}_t^{(k)}\right) + \lambda\, \mathbf{u}^{\top}_t\Sigma^{-1} \boldsymbol{\epsilon}_t^{(   k)} \right],
\label{eq:mppi_cost}
\end{equation}
where $\tilde{\mathbf{u}}_t^{(k)} = \mathbf{u}_t + \boldsymbol{\epsilon}_t^{(k)}$. The nominal control sequence is updated at each iteration as the weighted sum of the sampled perturbations:
\begin{align}
    w^{(k)} &= \frac{\exp\!\left(-\frac{1}{\lambda}\left(S^{(k)} - \rho\right)\right)}{\sum_{j=1}^{K} \exp\!\left(-\frac{1}{\lambda}\left(S^{(j)} - \rho\right)\right)}, \\
    \mathbf{u}_t &\leftarrow \mathbf{u}_t + \sum_{k=1}^{K} w^{(k)}\, \boldsymbol{\epsilon}_t^{(k)},
\label{eq:mppi_update}
\end{align}
where $\lambda > 0$ is a temperature parameter and $\rho = \min_k S^{(k)}$. This shifts the nominal trajectory toward lower-cost regions of the state space. The previous solution is used to warm-start the optimisation at each subsequent iteration.

% In typical MPC, the objective is formulated as
%
% \begin{equation}
%     J = \sum_{t=1}^{T} \gamma^t \left( C^t \right) + \text{terminal cost}
% \end{equation}
%
% from the sequence of optimal control inputs .., the inputs are applied in a receding horizon fashion.
%
% many Monte Carlo rollouts are generated in parallel and costs associated with trajectory is denoted by

\section{Our Approach}
\label{sec:approach}

Our framework for search and capture is illustrated in \autoref{fig:flow_diag}. We use a dynamic occupancy grid to represent the belief state of the environment. The hybrid planning framework operates in two modes: a global informative planning mode for exploration and target tracking, and an interception mode for physical capture. The vehicle switches to interception mode when a target is detected within a range threshold $R_s$, reverting to the informative planner after a successful or failed capture attempt. Our key contribution is an MPPI-based informative planner that couples target drift predictions with exploration and trajectory feasibility objectives, supporting effective search and capture of dynamic targets. In the interception mode, we use pure pursuit guidance based on the target depth estimate.

\subsection{Mapping}

We use a forward-facing camera to detect and localise targets of interest, and a dynamic occupancy grid $\mathcal{M}$ to represent the positions of targets in a global map, following the approach of Ramkumar Sudha et al. \cite{sanjeev2026informative}. For predictions, the wind measurements are used to predict the motion of the occupied positions in the map, and we assume targets to drift at $2$-$4\%$ of the wind speed based on the leeway drift approximation \cite{wagner2022winds}. Higher-fidelity models could improve prediction accuracy but at added computational cost, and therefore we use a lightweight approximation that aids our real-time planning requirements.
We model the field of view of the camera as an arc with an angular field of view corresponding to the camera intrinsics, and the radial field of view is limited to a distance $R_d$. For the occupancy grid, we use distance-based inverse sensor models with confidence decaying exponentially with distance, similar to \cite{moon2022tigris, popovic2020informative}. 
% However, the approach in \cite{sanjeev2026informative} does not explicitlyconsider the time-varying uncertainty in the map due to the dynamic nature of the environment. Therefore, in addition to a dynamic occupancy grid, we maintain a visit score grid $\mathcal{V}$ to capture the time-varying uncertainty in the map. 

To account for time-varying uncertainty in the map, in addition to the occupancy grid, we maintain a visit score grid $\mathcal{V} \in [0,1]$. Previously observed cells get progressively less confident over time, with $0$ representing currently observed cells and more uncertain cells denoted by higher values of $\mathcal{V}$. 
Let $\mathcal{F}^t \subseteq \mathcal{M}$ denote the set of grid cells inside the field of view at time $t$. The grid is updated as follows:
\begin{equation}
    v_{i, j} = 
    \begin{cases}
        0 & (i,j) \in \mathcal{F}^t \\
        \min\!\bigl(v_{i, j} + \eta\,\Delta t_o,\; 1\bigr) & (i,j) \notin \mathcal{F}^t,
    \end{cases}
\end{equation}
where $\eta$ is a decay rate hyperparameter and $\Delta t_o$ is the time step between successive observations. 

\subsection{Informative trajectory planning with MPPI}
\label{sec:planning}

% \textcolor{red}{Structure the planning part as a separate section instead? sec III-mapping and sec IV-planning approach?}

% In this section, we describe the formulation of the proposed spatiotemporal informative planning approach based on MPPI.
In this section, we describe the formulation of our MPPI-based spatiotemporal informative planning approach that enables planning continuous trajectories for search and capture in environments with time-varying target distributions.

\subsubsection{System dynamics}
% \noindent\textbf{Vehicle kinematics:} 
We use the vehicle kinematics as approximate system dynamics for the MPPI planner. The state $\mathbf{x}_t = [x_t,\, y_t,\, \psi_t]^\top$ consists of the position $(x_t, y_t)$ and the heading $\psi_t$ of the vehicle. The control input is $\mathbf{u} = [u,\, r]^\top$, consisting of the surge speed $u$ and the yaw rate $r$. The discrete-time kinematic model is defined as:

% \begin{equation}
%     \mathbf{x}^{t+1} = \underbrace{\begin{bmatrix} 1 & 0 & 0 \\ 0 & 1 & 0 \\ 0 & 0 & 1 \end{bmatrix}}_{A}\mathbf{x}^t +
%     \underbrace{\begin{bmatrix} \cos\psi^{t+1} & 0 \\ \sin\psi^{t+1} & 0 \\ 0 & 1 \end{bmatrix}}_{B(\psi^t)}\mathbf{u}^t \,\Delta t_p
% \end{equation}

\begin{equation}
    \mathbf{x}_{t+1} = \mathbf{x}_t +
    {\begin{bmatrix} \cos\psi_{t} & 0 \\ \sin\psi_{t} & 0 \\ 0 & 1 \end{bmatrix}}\, \Delta t_s\,\mathbf{u}_t ,
\end{equation}
% where $\psi^{t} = \psi^{t-1} + r^{t-1} \Delta t$ and 
where $\Delta t_s$ is the planning time step used in the predictions. 
This model approximates the vehicle motion in the horizontal plane while remaining lightweight enough for evaluating hundreds of parallel rollouts at high frequency for the MPPI planner.
For each rollout, the occupancy grid updates are forward propagated for sampled poses on the trajectory to calculate the planning costs. This requires estimating the field of view along poses on each sample trajectory.

% The global planner considers exploring regions with high uncertainty and tracking moving targets with the ASV.  is used as the global planner. The reward consists of terms for reducing uncertainty in the map, tracking targets (by reducing distance to the targets), collision costs, and path cost (for MPPI). This steers the ASV to get closer to targets while reducing uncertainty in the map with feasible trajectories.
% For capture/interception of targets, the ASV needs to approach the targets with almost zero bearing while reducing distance between them and the ASV.  This requires predicting both the target’s and the ASV’s motion over short horizons. The global planner, however, considers longer time horizons for mapping/exploring. The latter is unsuitable for the interception stage as precise control is required for successful capture. Therefore, once a target is within a specified range, we switch to intercept/capture mode. We use pure pursuit/constant bearing for guidance of the ASV in the capture stage.

\subsubsection{Planning costs}
We design costs for the MPPI planner that balance between exploration and target interception while ensuring safe, feasible trajectories. The stage cost at a single time-step $C$ is computed as:
\begin{equation}
    C = a_1 C_{e} + a_2 C_{t} + a_3 C_c + a_4 C_p,
\label{eq:planning_utility}
\end{equation}
where $C_e$ is the exploration cost to reduce uncertainty along the trajectory, $C_t$ is the tracking cost that enables approaching the targets, $C_c$ is the collision and boundary cost, and $C_p$ is the path cost. The coefficients $a_1,\, a_2,\, a_3,\, a_4$ are hyperparameters that we tune empirically to attain the desired behaviour in terms of task performance and trajectory smoothness.
% to trade off between the different objectives and achieve the desired behaviour. 

% \begin{multline}
% C_e &= \\ 
% C_t &= \\
% C_c &= \\
% \end{multline}

% While MPC-based approaches are commonly used in . Specifically, we make two modifications to make MPPI suitable for informative planning.
%%%%%%%%%%%%
% We design costs for the MPPI that balances between exploration, target tracking and interception, dynamic feasibility, and collision costs. The total trajectory cost $J$ is the discounted sum of per-timestep costs over the planning horizon $T$:

% \begin{equation}
%     J = \sum_{t=1}^{T} \gamma^t \left( a_1 C_{e}^t + a_2 C_{t}^t + a_3 C_{p}^t + a_4 C_{c}^t \right)
% \end{equation}

% where $\gamma \in (0,1]$ is a discount factor, $C_e^t$ is the exploration cost that rewards reducing unvisited regions along the trajectory, $C_t^t$ is the tracking cost that promotes approaching target vicinities. $C_c^t$ is the collision cost, and $C_p^t$  is the path cost. 

% \textbf{Exploration cost}
The exploration cost reduces uncertainty in the map. It rewards trajectories that observe cells with high scores in the visit score grid. 
For each sampled trajectory, updates to the visit score grid $\mathcal{V}$ are simulated through parallelised computations. 
It is computed by summing the predicted visit scores $\mathcal{V}$ over cells in the field of view along each sampled trajectory, normalised by grid dimensions $H\,\times\,W$:
\begin{equation}
    C_e = -\frac{1}{HW}\sum_{(i,j) \in \mathcal{F}^t} v_{i,j}.
\end{equation}
% \textbf{Target interception cost} 
Target tracking cost rewards trajectories that reduce the distance and bearing to targets. Spatiotemporal target predictions $\hat{p}_{i,j}$ over the planning horizon are obtained with a modified UNet as described in \cite{sanjeev2026informative}. The cost is defined as:

\begin{equation}
    C_t = -\sum_{(i,j) \in \mathcal{F}^t} \hat{p}_{i,j}\, e^{-\!\bigl(c_1\,d_{i,j}^2 + c_2\,\theta_{i,j}^{2}\bigr)},
\label{eq:track_cost}
\end{equation}
where $d_{i,j}$ and $\theta_{i,j}$ are the distance and the bearing of an arbitrary cell relative to the vehicle pose, and $c_1, c_2$ are parameters that affect the strength of the virtual potential field around a target that pulls the vehicle towards it.

% \textbf{Path cost} 
The path cost reduces unnecessary turns by penalising large yaw rates and ensures that the vehicle moves with the desired speed by penalising deviations from the desired surge speed. It is calculated as:
% using normalised control inputs $\bar{u}^t = (u^t - u_{\min})/(u_{\max} - u_{\min})$ and $\bar{r}^t = |r^t|/r_{\max}$:
\begin{equation}
    C_p = \bar{u}_t^2 + \bar{r}_t^2,
\end{equation}
where $\bar{u}_t = (u_t - u_{d})/(u_{\max} - u_{\min})$ and $\bar{r}_t = |r_t|/r_{\max}$ are normalised surge speed and yaw rate, and ${u}_d$ is the desired surge speed.

% \textbf{Collision cost}
Collision cost ensures that the vehicle stays in the search area and avoids collisions. A penalty is applied as a soft barrier function on the minimum distance to any boundary of the navigable region:
\begin{equation}
    C_c = \log\!\left(1 + e^{\,d_s - d_{\min}}\right),
\end{equation}
where $d_{\min}$ is the minimum distance to any boundary of the navigable region and $d_s$ is a safety distance parameter.

% We tune the costs empirically

\subsubsection{Time-correlated noise}

% \begin{figure}
%     \subfloat{
%         \includegraphics[width=0.495\linewidth]{images/iid_gaussian_distribution.png}}
%     % \hfil
%     \subfloat{
%         \includegraphics[width=0.495\linewidth]{images/ou_distribution.png}}
%     \caption{Sampling trajectory distributions with (a) random Gaussian noise and (b) time-correlated OU process noise. The latter generates a more diverse trajectory set. The mean trajectory is shown in red and the sampled trajectories are in blue.}
%     \label{fig:trajectory_distribution}
% \end{figure}

\begin{figure}[!htbp]
    \centering
    \includegraphics[width=0.9\linewidth]{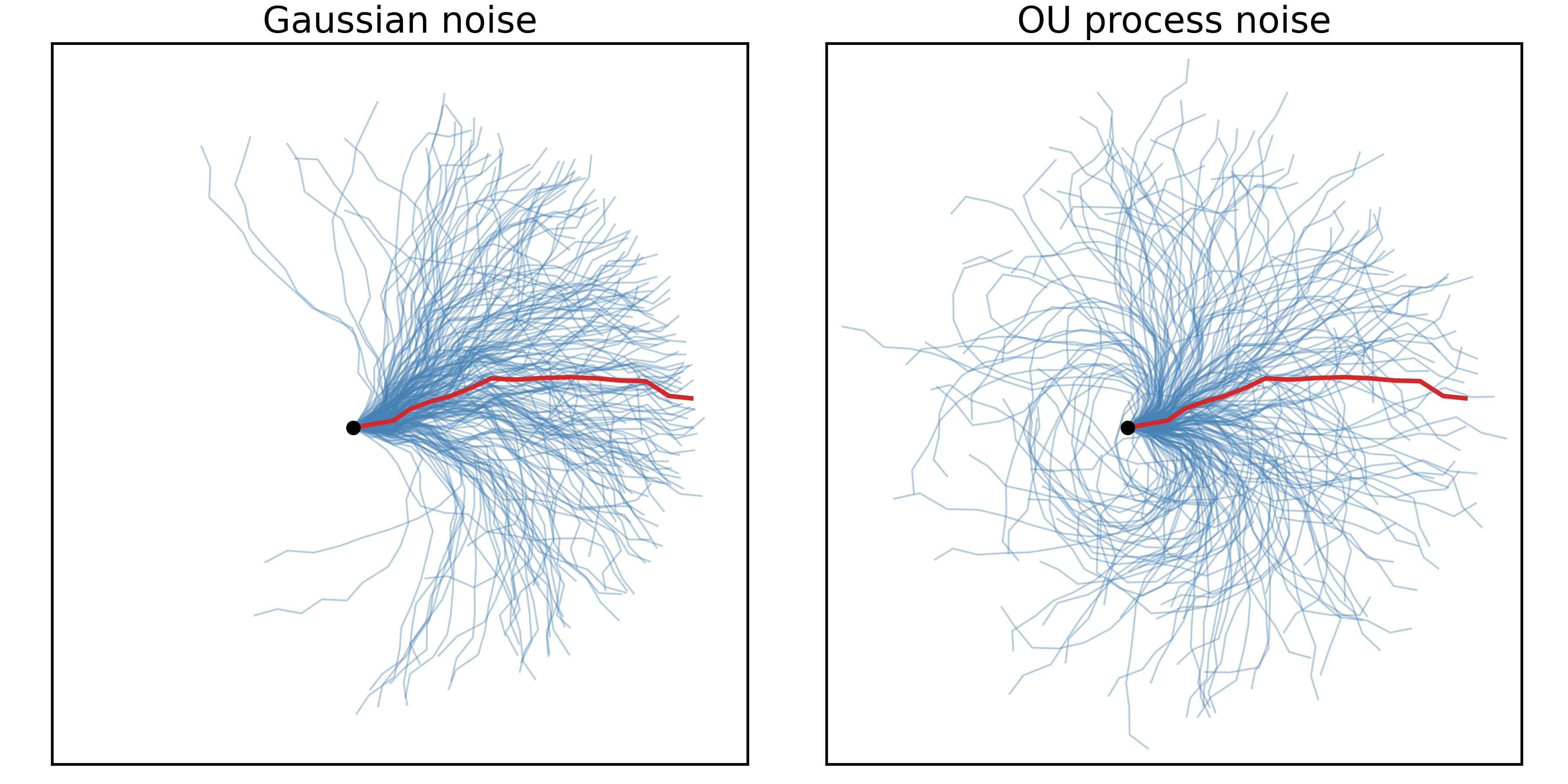}
    \caption{Sampling trajectory distributions with random Gaussian noise and time-correlated OU process noise visualised. The latter generates a more diverse trajectory set, facilitating better exploration. The mean trajectory is shown in red, and the sampled trajectories are in blue.}
    \label{fig:trajectory_distribution}
\end{figure}

Standard MPPI samples independent Gaussian noise at each time step, which can produce jerky control sequences and limit trajectory diversity. Incorporating structured sampling approaches such as Halton sequences has been shown to improve action space exploration in MPPI \cite{bhardwaj2022storm}. In a similar spirit, we use a discrete-time Ornstein-Uhlenbeck (OU) process to generate temporally correlated perturbations, to produce smoother, more diverse rollouts, as seen in \autoref{fig:trajectory_distribution}. The OU noise for sample $k$ is generated recursively over the planning horizon as:
\begin{align}
    \epsilon_t^{(k)} &= \alpha\, \epsilon_{t-1}^{(k)} + \sqrt{1 - \alpha^2}\; \delta_t^{(k)}, \quad \delta_t^{(k)} \sim \mathcal{N}(\mathbf{0}, \Sigma),
\end{align}
where $\alpha \in (0, 1)$ controls the temporal correlation, 
% with larger $\alpha$ producing smoother noise, 
and $\delta_t^{(k)}$ is uncorrelated Gaussian noise as in standard MPPI. \autoref{fig:trajectory_distribution} compares the sampling trajectory distributions generated with our time-correlated noise and random Gaussian noise. We omit the noise cost (see \eqref{eq:mppi_cost}), as penalising added noise inhibits exploration of the state space that is required to better satisfy our objectives. 
% as we find that this produces better trajectories for our objectives, since it requires better exploration of the state space and the noise cost inhibits this by penalising added noise.

\subsubsection{Warm-starting the optimisation}

% We use the MPPI planner to directly output guidance-level commands, i.e., the desired surge speed $u$ and yaw-rate $r$, to the low-level controllers. However, in the dynamics for planning, we use longer time-steps to reason over longer planning horizons without excessive computational cost. 
% Each MPPI iteration is warm-started with the previous solution. In standard MPPI, the nominal sequence is shifted forward by one step each iteration, discarding the already applied control input. 
We perform optimisation and apply control inputs at time steps of $\Delta t_c$. However, we use longer time-steps ($\Delta t_s$) for the system dynamics in planning to reason over longer horizons, since covering such horizons at the finer step $ \Delta t_c$ would make the computation prohibitive for real-time planning. Each MPPI iteration is warm-started with the previous solution. In standard MPPI, the nominal sequence is shifted forward by one step each iteration, discarding the already applied control input. Here we shift the nominal trajectory by one prediction step only once every $\Delta t_s/\Delta t_c$ iterations and reuse it unshifted otherwise, keeping the warm-start aligned with the coarser discretisation of the system dynamics. Moreover, the MPPI optimisation is still executed at the desired control frequency, i.e., $1/\Delta t_c$ Hz, even when the vehicle is in interception mode, to ensure that there are no lags in warm-starting the optimisation after switching between exploration and interception modes.

\subsection{Target interception}

For the physical capture of targets, we use pure pursuit guidance to guide the vehicle to intercept targets. The depth measurements from the camera are used to estimate the target position during interception, as the mapping resolution and time step of the global MPPI planner are too coarse for the precise terminal guidance required during interception. We compute the desired yaw rate for the vehicle as:
\begin{equation}
    r = k\, \psi_d,
\end{equation}
where $k$ is a proportional gain parameter, and $\psi_d$ is the desired relative bearing to a target, obtained from depth measurements. For a target with estimated position $\mathbf{p_t}=(p_x, p_y, p_z)$ in the vehicle coordinate frame, $\psi_d$ is computed as  $ \psi_d = \tan^{-1}(p_y/p_x)$. In interception mode, only the nearest target is pursued when multiple targets are in view.
% The ASV switches back to the global MPPI planner once the target leaves the field of view resulting in a successful or failed interception. The surge speed is kept at the desired speed $u_d$ throughout the interception stage. 
%%%%%%%%%%%
\section{Experimental Evaluation}

Our experiments are designed to (i) validate the claim that our MPPI planner outperforms other planning baselines with greedy and exploratory behaviours; (ii) evaluate the effectiveness of our proposed cost design for the MPPI planner. Simulation experiments are performed in a high-fidelity Gazebo environment to validate these two claims. We also conduct field trials to validate our framework for real-world scenarios with a real ASV.

\subsection{Experimental setup}

We perform simulation experiments with the Virtual Robot-X (VRX) simulator \cite{bingham2019toward}. Wind is modelled as a stochastic first-order Gauss-Markov process, and wave effects are also simulated to reproduce realistic sea conditions. A search area of $100$ m $\times\,100$ m is considered, and objects of $0.4$ m in diameter are generated to mimic floating litter that freely drift due to the environmental forces.
We create three scenarios for evaluation, with $10$ targets in two, and $20$ in the third. A single trial lasts $180$ s, with target spawn positions fixed per scenario. The environmental forces are stochastic, with moderate-to-high mean wind speeds of $8–12$\, m/s. We perform Monte Carlo simulations by repeating each scenario for 10 runs for the evaluations. 
% We consider three scenarios with $10$ targets in the first two and $20$ in scenario $3$, under moderate-to-high mean wind speeds of $8–12$\, m/s.

For mapping, we use a grid resolution of $1$ m. The visit score and occupancy grids are both updated at $5$ Hz. For the MPPI planner, we use a horizon of $30$ s with time steps of $2$ s. We generate velocity commands by replanning at $10$ Hz to generate smooth trajectories for the vehicle, and use $150$ rollouts at each step. We set the coefficients $a_1=1,\,a_2=0.45,\,a_3=0.05$ and $a_4=0.1$ after empirically tuning them. Simulation experiments are performed on a desktop with an AMD Ryzen 2950X CPU and Nvidia RTX 2080 Ti GPU.

For evaluation, we choose the following metrics: (i) the total number of successful interceptions (N), a measure of task completion; (ii) time taken to capture half the targets in the environment ($T_{1/2}$), only reported for runs with $N$ exceeding this threshold on average; and (iii) the reduction in the Shannon entropy ($\Delta H$) of the occupancy grid at the end of mission time, which is indicative of the area explored.
% (iii) the mean value of $\mathcal{V}$ during the mission time ($\overline{\mathcal{V}}$) which gives an average measure of uncertainty in the map.

\subsection{Comparison with planning baselines}

\begin{table*}[!ht]
    \centering
    \caption{Number of successful captures (N) and entropy reduction ($\Delta H$) for different planning strategies.}
    \begin{tabular}{|l|ccc|ccc|ccc|}
    \hline
     & \multicolumn{3}{c|}{Scenario 1} 
    & \multicolumn{3}{c|}{Scenario 2}
        & \multicolumn{3}{c|}{Scenario 3} \\
    \textbf{Strategy} & \textbf{N} & \textbf{$T_{1/2}$}& \textbf{$\Delta$ H} & \textbf{N} & \textbf{$T_{1/2}$} & \textbf{$\Delta$ H} & \textbf{N} & \textbf{$T_{1/2}$} & \textbf{$\Delta$ H} \\
    \hline
    % Coverage  & $1.3\pm0.675$ & & $0.236\pm0.008$ & $1.6\pm0.699$ & - & $0.246\pm0.003$ & $1.9\pm1.286$ & - & $0.24\pm0.007$\\
    % Coverage-greedy  & $2.3 \pm 0.949$ & & $0.156 \pm 0.026$ & $3.7 \pm 0.949$ & - & $0.232 \pm 0.032$ & $8.2 \pm 1.32$ & - & $0.260 \pm 0.026$ \\
    % Frontier  & $3\pm0.816$ & & $\mathbf{0.347 \pm0.028}$ & $2.3\pm0.948$ & - & $\mathbf{0.353\pm0.043}$ & $2.3\pm 2.002$ & - & $0.334\pm0.052$ \\
    % Frontier-greedy  & $8.6 \pm 1.35$ & $102 \pm 23.94$ & $0.323 \pm 0.029$ & $5.6 \pm 0.966$ & $134.4 \pm 15.06$& $0.333 \pm 0.023$ & $11.2 \pm 2.30$ & $158.7 \pm 24.6$ & $0.307 \pm 0.035$ \\
    % RH  & $5.7 \pm 0.483$ & $108 \pm 20.44$ & $0.307 \pm 0.017$ & $5.7 \pm 0.949$ & $132 \pm 12.25$& $0.303 \pm 0.018$ & $11.4 \pm 2.17$ & $153.4 \pm 10.69$ & $0.311 \pm 0.018$ \\
    % RH-greedy  & $8.4 \pm 0.516$ & $84 \pm 5.16$ & $0.335 \pm 0.024$ & $7.2 \pm 0.422$ & $119 \pm 3.33$ & $0.332 \pm 0.011$ & $13.6 \pm 1.07$ & $128.8 \pm 11.91 $ & $0.323 \pm 0.016$ \\
    % Ours  & $9.4 \pm 0.516$ & $\mathbf{68 \pm 7.89}$& $0.319 \pm 0.022$ & $\mathbf{8.5 \pm 0.707}$ & $\mathbf{110 \pm 9.28}$ & $0.333 \pm 0.038$ & $\mathbf{15.6 \pm 1.43}$ & $\mathbf{121.7 \pm 13.29}$ & $\mathbf{0.344 \pm 0.009}$ \\
    % Ours-greedy  & $\mathbf{9.7 \pm 0.949}$ & $80 \pm 0$& $0.331 \pm 0.01$ & $7.7 \pm 0.483$ & $116 \pm 5 $ & $0.326 \pm 0.014$ & $13.9 \pm 0.994$ & $122.9 \pm 10.4$ & $0.324 \pm 0.012$ \\

    Coverage  & $1.3\pm0.7$ & - & $0.236\pm0.008$ & $1.6\pm0.7$ & - & $0.246\pm0.003$ & $1.9\pm1.3$ & - & $0.24\pm0.007$\\
    Coverage-greedy  & $2.3 \pm 0.9$ & - & $0.156 \pm 0.026$ & $3.7 \pm 0.9$ & - & $0.232 \pm 0.032$ & $8.2 \pm 1.3$ & - & $0.260 \pm 0.026$ \\
    Frontier  & $3.0\pm0.8$ & - & $\mathbf{0.347 \pm0.028}$ & $2.3\pm0.9$ & - & $\mathbf{0.353\pm0.043}$ & $2.3\pm 2.0$ & - & $0.334\pm0.052$ \\
    Frontier-greedy  & $8.6 \pm 1.3$ & $102 \pm 24$ & $0.323 \pm 0.029$ & $5.6 \pm 1.0$ & $134 \pm 15$& $0.333 \pm 0.023$ & $11.2 \pm 2.3$ & $159 \pm 25$ & $0.307 \pm 0.035$ \\
    RH  & $5.7 \pm 0.5$ & $108 \pm 20$ & $0.307 \pm 0.017$ & $5.7 \pm 0.9$ & $132 \pm 12$& $0.303 \pm 0.018$ & $11.4 \pm 2.2$ & $153 \pm 11$ & $0.311 \pm 0.018$ \\
    RH-greedy  & $8.4 \pm 0.5$ & $84 \pm 5$ & $0.335 \pm 0.024$ & $7.2 \pm 0.4$ & $119 \pm 3$ & $0.332 \pm 0.011$ & $13.6 \pm 1.1$ & $129 \pm 12 $ & $0.323 \pm 0.016$ \\
    Ours  & $9.4 \pm 0.5$ & $\mathbf{68 \pm 8}$& $0.319 \pm 0.022$ & $\mathbf{8.5 \pm 0.7}$ & $\mathbf{110 \pm 9}$ & $0.333 \pm 0.038$ & $\mathbf{15.6 \pm 1.4}$ & $\mathbf{121 \pm 13}$ & $\mathbf{0.344 \pm 0.009}$ \\
    Ours-greedy  & $\mathbf{9.7 \pm 0.9}$ & $80 \pm 0$& $0.331 \pm 0.01$ & $7.7 \pm 0.5$ & $116 \pm 5 $ & $0.326 \pm 0.014$ & $13.9 \pm 1.0$ & $123 \pm 10$ & $0.324 \pm 0.012$ \\
    
        \hline
    \end{tabular}
    \label{tab:sim_results}
\end{table*}

% To evaluate our planner in its performance for search and capture of , is compared against other decision-making baselines
The proposed MPPI informative planner is compared against other decision-making strategies with our planning framework to evaluate its performance in terms of the number of captures in a given mission time in environments with no prior information about the targets.
% in terms of the number of successful captures within a limited mission time.
% We perform simulation experiments with an ASV using the Virtual Robot-X (VRX) simulator \cite{bingham2019toward}. A search area of $100$ m $\times 100$ m is considered, and a single trial lasts for a maximum duration of $150$ s or until all targets are captured.
We choose the following planning baselines:

\begin{itemize}
    % \item Our MPPI planner with the entire framework, utilising the planning cost described in \eqref{eq:planning_utility}. 
    \item A receding horizon (RH) planner that selects the action with lowest cost from a finite set of discretised control behaviours \cite{johansen2016ship}, with each trajectory corresponding to constant speed and yaw rate. It uses the same planning costs and horizon as our MPPI planner to fairly isolate the effect of using continuous trajectory optimisation over discrete action spaces. 
    \item A short-sighted frontier-based exploration strategy that selects a single waypoint for exploration based on the nearest frontiers approach \cite{yamauchi1997frontier}. 
    \item Coverage planner with lawnmower motions, a non-adaptive strategy commonly used for monitoring \cite{galceran2013survey}.
\end{itemize}

For target capture, we consider two strategies to isolate the effect of the exploration strategy from the target interception behaviour. First, a greedy target chasing strategy where the vehicle switches to interception with pure pursuit guidance if there is a target in field of view ($R_s=R_d=35$\,m). In the second strategy, we consider a different switching condition where we switch to interception mode only once the vehicle is within close proximity to a target ($R_s=8$\,m).
% The results are summarised in \autoref{tab:sim_results}.

From \autoref{tab:sim_results}, we observe that for the baseline planners, greedy target chasing consistently results in more captures when compared to switching to interception mode only when the targets are in close vicinity. This is because none of the baselines optimise continuous trajectories. Coverage and frontier strategies do not consider target tracking along the trajectories.
% This can be attributed to the frontier planner being myopic and not considering target interception explicitly in the planning objective. While the RH planner uses the same planning cost, as the MPPI planner, it does not plan trajectories in continuous action space, resulting in it being not as efficient as the MPPI planner in task completion. 
For our MPPI planner, however, the switching condition has little impact, i.e., greedy switching leads to $3.2\%$ more captures in scenario $1$ and fewer captures in scenarios $2$ and $3$ by $9.4\%$ and $10.9\%$ respectively. This can be attributed to the tracking cost \eqref{eq:track_cost} of our planning objective \eqref{eq:planning_utility}, which plans trajectories that steer the vehicle to the vicinity of targets, reducing reliance on pure pursuit guidance to capture targets.
% while in the global exploration mode. 
% The transition to pure pursuit thus occurs naturally as the vehicle approaches the target, making the explicit switching criterion less consequential.

\begin{figure}[!tbp]
    \centering
    \includegraphics[width=0.96\linewidth]{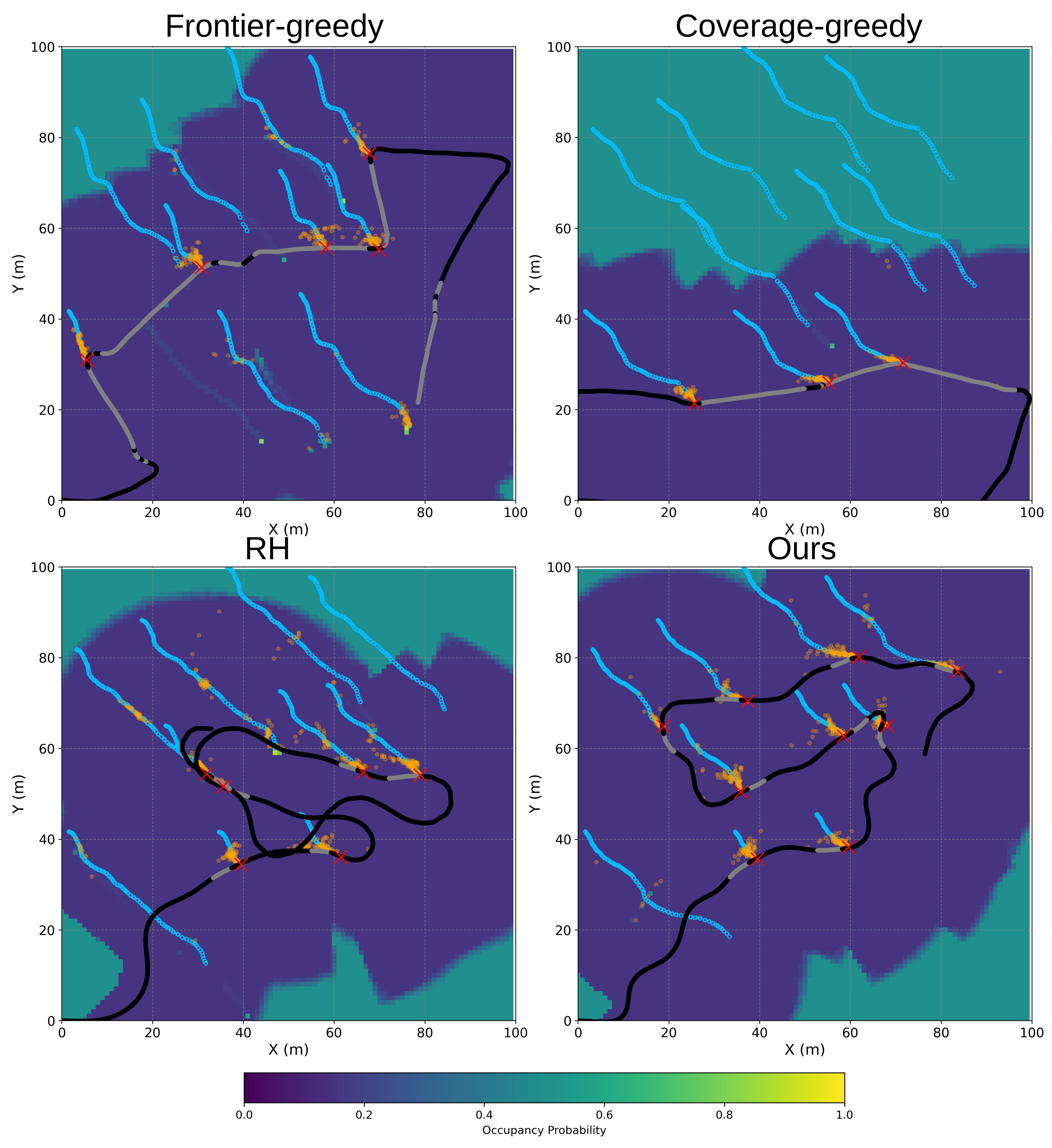}
    \caption{A qualitative analysis of the paths followed for a single stochastic scenario. The path segments in grey and black denote portions when the vehicle is in interception and exploration modes, respectively, and the red crosses denote successful interceptions. The blue trails denote ground truth positions of the targets and the orange dots represent the target detections. 
    % The frontier and coverage planners rely on greedy target captures, as they both do not consider target capture explicitly in the planning utility. While the RH and our MPPI planner both use the same utility, our planner captures more targets as it directly optimises continuous trajectories as opposed to planning over discrete actions.
    }
    \label{fig:path_plots}
\end{figure}

The coverage planner performs the worst in terms of captures, as it is not an adaptive strategy. Frontier exploration records the highest entropy reduction in $2$ of the $3$ scenarios, showing that it is effective for spatial exploration. However, this does not lead to more captures, as it is myopic and does not consider target tracking in the planning objective. Meanwhile, the RH and MPPI-based planners that use our planning costs outperform the frontier exploration strategy, highlighting the advantage of our cost design. Although the RH planner uses the same planning cost and horizon as the MPPI planner, it only plans over a discretised action space, thereby limiting the trajectory diversity, leading to fewer captures.
% We notice that the RH planner with greedy interception is the best performing of the baselines, which indicates the effectiveness of our proposed planning objective. 
Even with greedy chasing, the RH planner leads to $\approx10.6,15.3$ and $12.8\%$ fewer captures in the $3$ scenarios, respectively, as compared to our MPPI planner, demonstrating the advantage of optimising trajectories in continuous space with our planner. The MPPI planner with non-greedy interception records the lowest $T_{1/2}$ in all scenarios, further supporting our planner cost design. We also plot the paths followed by the vehicle with each planning strategy in \autoref{fig:path_plots}.

% \subsection{Ablation experiments?}

% We also study the effect of our time-correlated noise strategy in the MPPI sampling distribution. It is compared against the and evaluate the control effort and the mean interceptions.

% \begin{table}[h]
%     \centering
%     \caption{Number of interceptions and control effort compared for }
%     \label{tab:ablation_results}
%     \begin{tabular}{|l|c|c|}
%         \hline
%         \textbf{Planner} & \textbf{Avg. interceptions} & \textbf{Control effort} \\
%         \hline
%         & -- & -- \\
%         & -- & -- \\
%         & -- & -- \\
%         \hline
%     \end{tabular}
% \end{table}

% We find that the MPPI planner with the time-correlated noise reduces control effort while maintaining task success rates...

\subsection{Ablation experiments}

\begin{figure*}[!ht]
\includegraphics[width=0.98\linewidth]{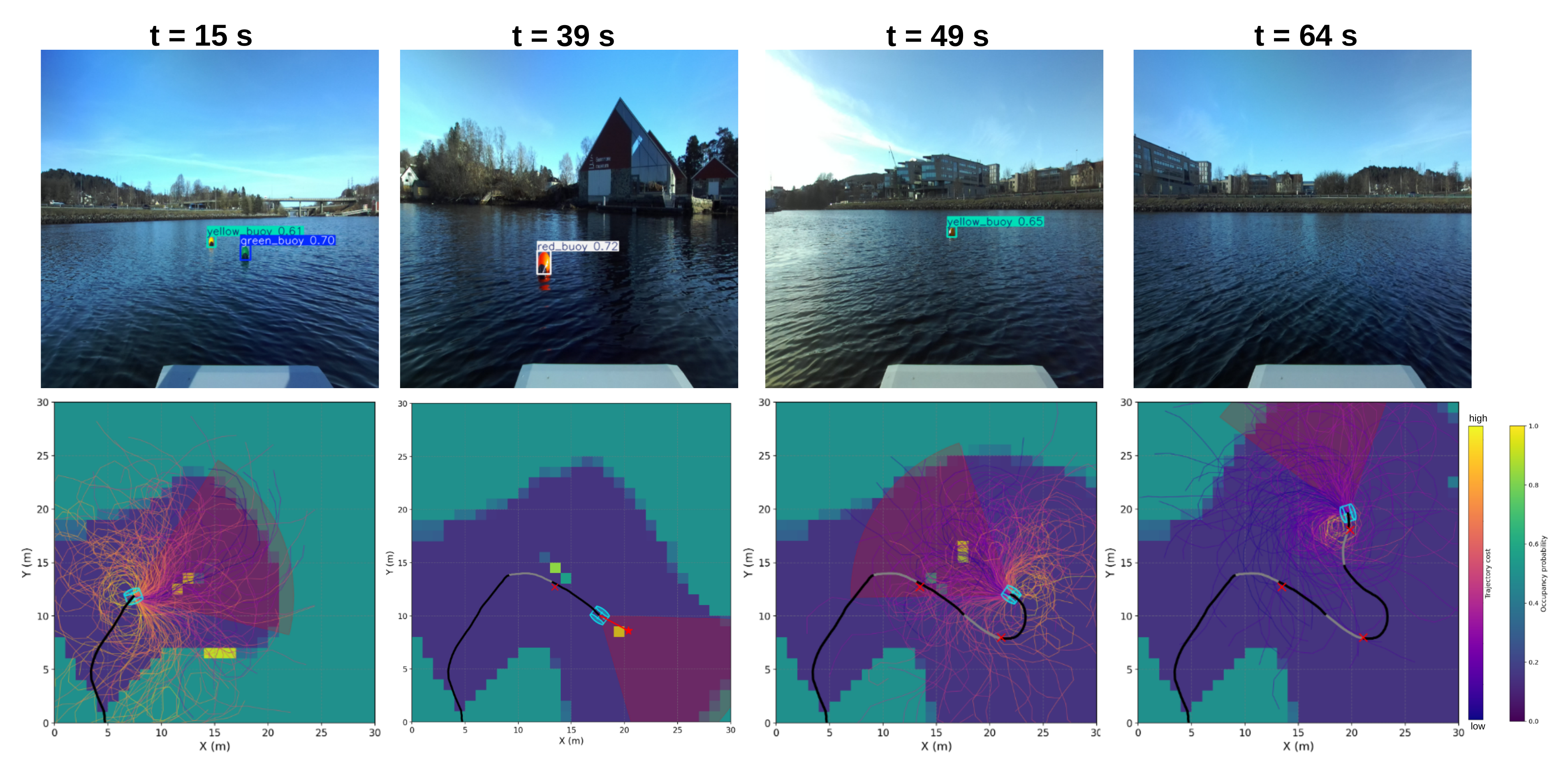}
\caption{Results from a field test showing the environment states at different time instants. At $t=15$\,s, the ASV has detected all 3 targets and tracks 2 targets. Later at $t=39$\,s, the ASV has captured 1 target and is pursuing the 2nd target, and at $49$\,s, the planner steers the vehicle into view of the final target after capturing the 2nd target. Finally, at $t=64$\,s, all 3 targets have been captured. The portions of the path followed by the vehicle in grey and black denote the segments it is in intercept and exploration modes, respectively. The red crosses denote locations of target interceptions. The sampled MPPI trajectories at the time instants are also shown with darker shades representing lower cost trajectories that better optimise the objective.}
\label{fig:field_tests}
\end{figure*}

To validate our cost design and study the effect of individual components of the planning cost in \eqref{eq:planning_utility}, we perform ablation experiments. Experiments from the previous section are repeated with our MPPI planner by setting the exploration, tracking, and path costs to zero, each in isolation. We also set the coefficients $c_1$ and $c_2$ in \eqref{eq:track_cost} to zero to further analyse the tracking cost design. For switching, we switch when in close vicinity to the target, as greedy target capture does not accurately reflect the trajectory quality.
% The results are summarised in \autoref{tab:ablation}.

\begin{table}[!htbp]
    \centering
    \caption{Ablation study of planning cost.}
    \begin{tabular}{|l|cc|cc|}
    \hline
     & \multicolumn{2}{c|}{Scenario 2} 
    & \multicolumn{2}{c|}{Scenario 3} \\
     & \textbf{N} & \textbf{$\Delta$ H}& \textbf{N} & \textbf{$\Delta$ H} \\
    \hline
    None  & $\mathbf{8.5 \pm 0.71}$ & $0.33 \pm 0.04$ & $\mathbf{15.6 \pm 1.43}$ & $0.34 \pm 0.01$\\
    $a_1 = 0$  & $8 \pm 0.94$ & $0.33 \pm 0.01$ & $14.7\pm 1.77$ & $0.34\pm0.02$\\
    $a_2 = 0$  & $2.4 \pm 0.84$ & $0.28 \pm 0.01$ & $6.6\pm2.06$ & $0.28\pm0.01$\\
    $a_4 = 0$  & $8.1\pm0.57$ & $\mathbf{0.34 \pm 0.01}$ & $14.3\pm2.36$ &  $\mathbf{0.35\pm0.02}$\\
    $a_1,a_4=0$  & $7.3 \pm 0.82$ & $0.34\pm 0.01$ & $11.8\pm2.35$ & $0.30\pm0.04$\\
    $c_1,c_2=0$ & $2.8 \pm 1.13$ & $0.28\pm0.03$ & $6.7\pm1.95$ & $0.27\pm0.03$\\
        \hline
    \end{tabular}
    \label{tab:ablation}
\end{table}

\autoref{tab:ablation} reveals that for the case with $a_2 = 0$, the captures degrade significantly, since the planner does not consider target drift predictions along the trajectories. This highlights that the tracking cost \eqref{eq:track_cost}, incorporating the target drift predictions, is the most important component as it steers the vehicle towards targets.
Removing the exploration cost ($a_1=0$) or the path cost ($a_4=0$) causes a marginal performance decline in $N$, but does not impact the reduction in entropy in the map $\Delta H$. This indicates that the path cost implicitly enables spatial coverage, leading to more target detections by producing smooth forward-moving trajectories. Meanwhile, removing both the path and exploration costs causes a larger decline in $N$, of up to $14.1\%$ and $24.4\%$ respectively in the two scenarios as compared to the nominal case.
Ablating the coefficients $c_1, c_2$ in \eqref{eq:track_cost} also causes a huge drop in $N$. This is due to the planner rewarding trajectories that keep targets in view without approaching their vicinities. This further highlights the role of the tracking cost in promoting target interceptions.
% This indicates that both the path and exploration costs support effective exploration, leading to more target detections. 
% When both are absent simultaneously, this regularisation is lost and task performance degrades more substantially.
% These results demonstrate the effectiveness of our proposed multiobjective cost for trajectory optimisation. 
% The tracking cost provides the dominant signal for exploitation, the exploration cost ensures that uncertain regions are visited, and the path cost improves both tracking and exploration by producing smoother trajectories. % -> remove this line?
These results validate our cost design for the MPPI planner and confirm that jointly optimising exploration and target interception over continuous trajectories enables effective search and capture.

\subsection{Field tests}

We conduct field experiments with a $1.2$ m long catamaran ASV (see \autoref{fig:fig1}) to validate our approach in real-world monitoring scenarios. It is equipped with an autopilot for state estimation and velocity control. The control commands output from the hybrid planner to the autopilot are the surge speed ($u$) and yaw rate ($r$). Our framework is implemented with ROS2 as middleware on a Jetson Orin NX on the ASV. 
A ZED 2i stereo camera is used for object detection, with a horizontal field of view of $72^\circ$ and a radial field of view of $15$ m. % -> remove this?
% For the capture of targets, we intercept targets using pure pursuit guidance and collect them with a net placed between the two hulls of the catamaran.
Targets are collected with a net placed between the two hulls of the catamaran.

We perform tests on a $900$ m$^2$ search area in a strait. Three buoy markers are used as targets as proxies for floating litter, that freely drift under the influence of moderate winds with mean speed of $6$\,m/s during the tests. The surge speed and yaw rates are limited to $u \in [0.3, 0.75]$ m/s and $r \in [-0.53, 0.53]$ rad/s for the MPPI planner. For switching to pure pursuit, we take $R_s = 3$ m. Perception and mapping updates are performed at $4$ Hz. The MPPI planner is executed at $10$ Hz using $150$ samples at every step. The entire framework uses $\approx3$ GB of RAM on the onboard Jetson Orin NX. One iteration using $150$ rollouts takes $\approx 0.08$ s on the embedded hardware, within the $0.1$ s control period. 
% Compute time does not grow with the number of targets as they are represented in the occupancy grid, but scales with the number of rollouts and horizon.

% Each trial lasts for a duration of $120$ s or until all cap. We perform $3$ trials and report the results in \autoref{tab:}. We also visualise the state of the map at various instants from the trials.
\begin{table}[h!]
    \centering
    \caption{Summary of the field results}
    % \small % Set the smaller font size here
        \begin{tabular}{|c|c|c|}
        \hline
        & \textbf{No. captures} & \textbf{Time taken} \\ \hline
            Trial 1 & $2/3$ & -- \\ \hline
            Trial 2 & $3/3$ & $95$ s \\ \hline
            Trial 3 & $3/3$ & $61$ s \\ \hline
            Trial 4 & $3/3$ & $49$ s \\ \hline
        \end{tabular}
\label{tab:field_trials}
\end{table}

To better assess the robustness of our approach, we perform four trials in the same settings with varying environmental conditions, and the results are summarised in \autoref{tab:field_trials}. All targets except one are captured across four trials. A target is missed in trial $1$ as it drifts out of the search area before being detected. Trial $2$ takes longer to complete than the other successful trials. This is due to inaccurate mapping, as the wind speed used in the mapping predictions is kept fixed due to the absence of real-time wind measurements. Results from a demonstrative trial are visualised in \autoref{fig:field_tests} with snapshots of the map state, path followed, and sampled trajectories showing how our planner steers the vehicle towards the moving targets without greedily chasing them. A video of the field trial is available in the attachment.
% \href{https://imgur.com/a/FQ4v0SK}{\footnotemark}.
% \footnotetext{Video from field trial at \url{https://imgur.com/a/FQ4v0SK}}

% Trial 2 takes longer to capture all targets. As seen in fig., a target is missed due to mapping inaccuracy but it is later . A reason for this is that during the field tests, the wind speed used in the mapping predictions is kept fixed due to being unable to fetch realtime wind measurements from a wind sensor. Results from a single field trial with snapshots of the environment states at different instants are visualised in \autoref{fig:field_tests}. Video attachements from the field trial will be made available at ..

\section{Conclusions and Future Work}

In this letter, we introduce a spatiotemporal informative planning approach for search and capture of drifting targets in marine environments with ASVs. A key aspect of our method is a planning formulation using MPPI to enable efficient search and capture by jointly reasoning over predicted target drift and search objectives, while planning continuous trajectories. Evaluations show that our planner captures up to $18\%$ more targets than the strongest baseline. The ablation studies further support the planner cost design and its associated costs. Field experiments demonstrate its validity in real-world scenarios. 

Our mapping approach relies on wind measurements to predict target drift, and therefore, the mapping accuracy degrades in the absence of reliable wind measurements. Therefore, investigating better estimation techniques for predicting target drift is an avenue for future work.
While our approach performs well with a single robot in unobstructed search areas, it can be challenging in large or cluttered spaces. Therefore, further work also includes multi-robot coordination for efficient monitoring. 
% Another direction for further work is human-in-the-loop monitoring with vision language action (VLA) models.

\addtolength{\textheight}{-12cm}   % This command serves to balance the column lengths

%%%%%%%%%%%%%%%%%%%%%%%%%%%%%%%%%%%%%%%%%%%%%%%%%%%%%%%%%%%%%%%%%%%%%%%%%%%%%%%%

%%%%%%%%%%%%%%%%%%%%%%%%%%%%%%%%%%%%%%%%%%%%%%%%%%%%%%%%%%%%%%%%%%%%%%%%%%%%%%%%



% \section*{ACKNOWLEDGMENT}

%%%%%%%%%%%%%%%%%%%%%%%%%%%%%%%%%%%%%%%%%%%%%%%%%%%%%%%%%%%%%%%%%%%%%%%%%%%%%%%%

% \pagebreak
\balance
\bibliographystyle{IEEEtran}
\bibliography{references}

\end{document}